\documentclass[sigconf,nonacm]{acmart}
\renewcommand\footnotetextcopyrightpermission[1]{}
\AtBeginDocument{%
  }

\setcopyright{acmlicensed}
\copyrightyear{2018}
\acmYear{2018}
\acmDOI{XXXXXXX.XXXXXXX}
\acmISBN{978-1-4503-XXXX-X/2018/06}

\renewcommand\footnotetextcopyrightpermission[1]{}

\begin{document}

\title{UESF-Bench: Benchmarking and Probing for Unified Embodied Seeking and Following}
\author{
\textbf{
Kun Yu$^{1}$,
Jianhua Yang$^{2}$,
Yixiang Chen$^{2}$,
Changwei Wang$^{3}$,
Hongyuan Yu$^{4}$,
Yan Huang$^{2}$,
Fushuo Huo$^{6}$,
Ya Jing$^{5}$,
Zhumin Chen$^{1}$,
Keji He$^{1}$
}
}

\affiliation{
\institution{
$^{1}$Shandong University \quad $^{2}$Institute of Automation, Chinese Academy of Sciences \quad $^{3}$Qilu University of Technology \quad$^{4}$The Multimedia Department, Xiaomi Inc\quad
$^{5}$Beijing University Of Technology \quad $^{6}$Hong Kong Polytechnic University
}
}







\renewcommand{\shortauthors}{Trovato et al.}

\begin{abstract}
Language-guided human following is an important capability for embodied agents, but existing benchmarks typically assume that the target person is visible at the start of an episode. This setting simplifies the problem and overlooks a more realistic requirement: an agent often needs to first find a language-described target and then persistently follow that target in a dynamic environment. While recent work has started to study human search, existing settings are typically evaluated in task-specific scenarios and often rely on stronger prior knowledge of the environment. Moreover, they usually treat searching and following as separate tasks and still lack a unified benchmark for systematic evaluation. To address these limitations, we introduce the Unified Embodied Seeking and Following Benchmark (UESF-Bench), a large-scale and diverse benchmark for embodied human seeking and following. The benchmark requires agents to handle semantic-guided exploration, reliable behavior switching and recovery, and delayed identity grounding. To this end, we propose SeekFollow-VLA, a vision-language-action framework with a task-driven routing mechanism for latent phase inference and transition modeling between seeking and following. Experimental results show that SeekFollow-VLA achieves clear improvements over both single-head and dual-head baselines across single-person and multi-person environments, establishing a baseline for unified embodied seek-and-follow.
\end{abstract}




\begin{teaserfigure}
  \centering
  \includegraphics[width=\textwidth]{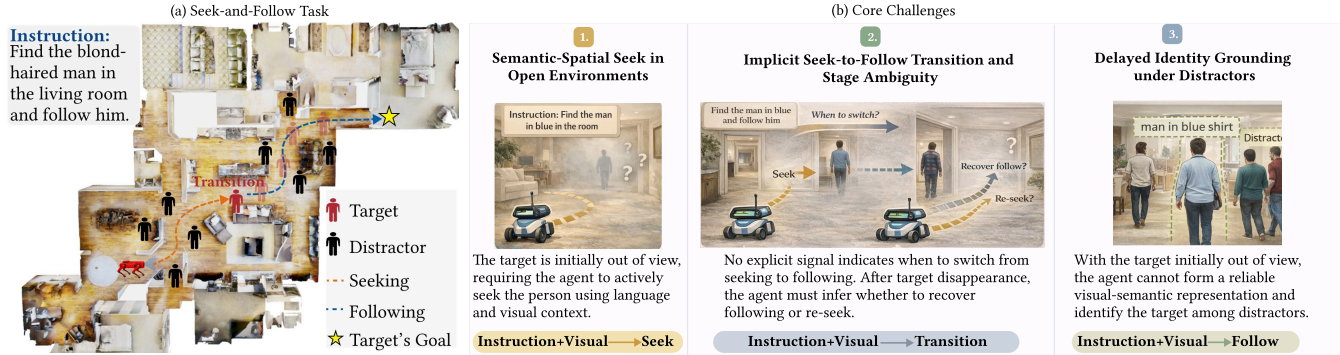}
  \caption{Overview of the embodied visual seek-and-follow task, its core challenges.}
  \label{fig:teaser}
\end{teaserfigure}


\maketitle

\section{Introduction}
Embodied human search~\cite{park2023socrates,fung2025mllm} and following~\cite{ye2025rpf,zhang2024uni,fung2025ldtrack,wang2025trackvla,liu2025trackvla++} are fundamental tasks for intelligent robots and have received increasing attention in the robotics community. In practical applications such as service robotics~\cite{lee2023developing,ostrowski2022mixed,mishra2024public}, elderly assistance~\cite{li2023mobile,fortuna2024personalizable,padmanabha2024voicepilot}, human-centered embodied interaction is often required. In such scenarios, an embodied agent is required to first locate a specific person and then seamlessly transition to persistent following in dynamic environments. This search-to-follow paradigm represents a more realistic form of embodied interaction, as it requires the agent not only to accomplish multiple subtasks, but also to maintain task continuity and behavioral consistency across different phases.

Existing language-guided human-following benchmark EVT-Bench~\cite{wang2025trackvla} has made important progress in embodied human intelligence. However, it relies on a key assumption that is often unrealistic in real-world scenarios, namely, the target person is initially visible. This assumption restricts the agent to learning localized perception-action mappings and consequently limits the capacity for modeling environmental structure and global spatial relationships. In real-world scenarios, the target is often not immediately observable within the agent’s initial field of view. For example, an instruction such as ``\emph{Find the blond-haired man wearing a fitted black leather jacket in the bedroom and follow him}.'' inherently requires an initial search phase, as the agent cannot begin following until the target has first been localized. This indicates that the following task alone is insufficient to capture the core functional requirements of real-world scenarios. This observation naturally leads to a more fundamental question: \emph{how can an agent autonomously localize a target in an unexplored environment}?

Although recent studies have begun to explore the human search problem, this line of research remains at an early stage. Existing methods~\cite{fung2025mllm,park2023socrates} are typically studied in simplified environments or under additional structured priors, such as explicit target cues and candidate regions. These settings limit their ability to systematically model complex search processes in open environments. Moreover, the lack of a unified evaluation benchmark makes it difficult to comprehensively assess an agent’s ability in semantic-guided exploration, spatial reasoning, and long-horizon decision-making. As a result, human-like search behavior, which relies on semantic and spatial commonsense for active target search in complex environments, remains insufficiently modeled and systematically evaluated. More importantly, simply treating search and follow as two independent tasks still fails to capture the continuity of real-world interaction. In practical scenarios, an agent must make decisions continuously under uncertainty: whether to keep searching, when to switch to following, and whether to resume search or continue following once the target is temporarily lost. This process fundamentally involves dynamic switching and coupling between different behavioral modes, rather than a static composition of separate tasks. Therefore, the core question is not how to evaluate searching and following independently, but how to unify them within a single framework and systematically assess the agent’s ability to switch between them during continuous embodied decision-making.

To fill this gap, we construct the Unified Embodied Seeking and Following Benchmark (UESF-Bench), a large-scale and diverse benchmark that covers a wide range of difficulty levels, resulting in a dataset of 1.43 million embodied seeking-and-following samples. UESF-Bench is designed to evaluate language-guided embodied seek-and-follow under a single fused instruction, covering both single-person and crowded multi-person environments. To the best of our knowledge, UESF-Bench is the first large-scale benchmark for unified language-guided human seeking and following in embodied settings.

Under this unified setting, the agent must autonomously complete the full process from target search to persistent following in an unseen environment based solely on a single fused natural-language instruction. As illustrated in Fig.~\ref{fig:teaser}, this setting introduces three core challenges. \emph{First}, semantic-spatial search in open environments requires the agent to actively infer potential target regions and perform goal-directed exploration from language descriptions and visual context when the target is initially out of view. \emph{Second}, implicit seek-to-follow transition and recovery ambiguity requires the agent not only to make a reliable transition from search to follow when the target is first observed, but also to determine whether to perform local recovery or resume search when the target is temporarily lost. \emph{Third}, the target is initially out of view, so the agent cannot establish a reliable visual-semantic representation of the true target at the beginning. It must therefore identify the correct person among distractors using only the language description before establishing persistent following and maintaining identity consistency.

To address these challenges, we propose SeekFollow-VLA, a vision-language-action framework for the unified embodied seek-and-follow task under a single fused instruction. The framework is designed to support semantic-guided target search in open environments, enable reliable behavior switching under implicit phase transitions, and handle delayed identity grounding under distractors. Specifically, SeekFollow-VLA adopts a task-driven routing design to reduce mode confusion between seeking and following behaviors. The proposed router enables dynamic task inference and transition modeling without relying on explicit task separation, thereby promoting more consistent behavior across different phases of the task. 
The contributions of this work can be summarized as follows:

\begin{itemize}
    \item We formulate embodied visual seek-and-follow as a unified embodied task where an agent must first seek a language-described target person who is initially out of camera view and then seamlessly transition to following that person in dynamic environments.
    \item We first construct UESF-Bench, a large-scale, diverse, and general benchmark for unified language-guided human seeking and following in embodied settings. UESF-Bench contains 1.43 million embodied visual seek-and-follow samples that cover a wide range of difficulty levels in both single-person and multi-person scenarios.

    \item We propose SeekFollow-VLA, a vision-language-action framework for unified embodied seek-and-follow under a single fused instruction. By explicitly modeling dynamic phase inference and seek-to-follow transition, SeekFollow-VLA provides a more effective solution to mode confusion between seeking and following behaviors.

    \item Extensive experiments in both single-person and multi-person environments demonstrate the effectiveness of the proposed method.
\end{itemize}

\section{Related Work}
\subsection{VLA Models for Embodied Tasks}
Recent advances in Vision-Language Models (VLMs)~\cite{liu2024deepseek,zeng2025glm,team2023gemini} have driven the development of Vision-Language-Action (VLA) models~\cite{zhong2025survey,sapkota2025vision,yu2025survey}, which augment pre-trained VLMs with action generation capabilities. VLA models integrate perception, language understanding, and action generation in an end-to-end framework, showing strong potential for general-purpose embodied control~\cite{kim2024openvla,black2024pi_0,zitkovich2023rt}. Although recent VLA models can support multiple tasks within a unified architecture, they largely depend on explicit task separation. Specifically, tasks are commonly specified either through one-to-one instruction--task mappings~\cite{liu2025trackvla++} or through the addition of special task tokens~\cite{zhang2024uni,wang2025trackvla} to distinguish predefined task types. As a result, these tasks are generally formulated and handled independently. In contrast, our task requires a single model to first seek a target person and then transition seamlessly to human following under a single fused instruction. This setting poses a new challenge for VLA design because the model must manage behaviorally distinct stages without any explicit task boundary. In particular, when the target person temporarily disappears from view, the model must infer from the evolving visual context alone whether it should continue following or return to seeking.

\begin{figure*}[t]
  \centering
  \includegraphics[width=0.93\linewidth]{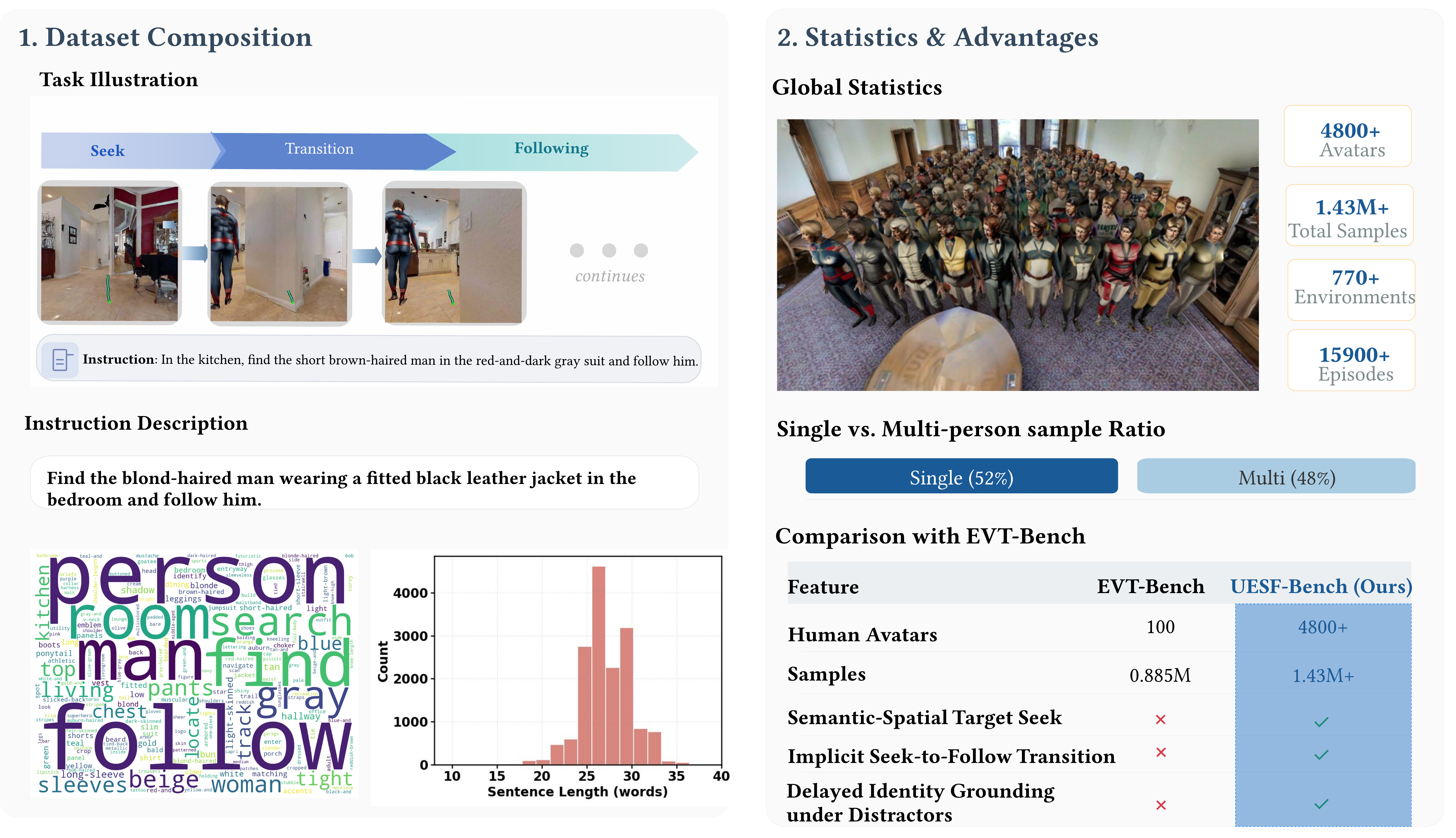}
  \caption{Overview and statistics of UESF-Bench. UESF-Bench contains 4800+ diverse humanoid avatars and 1.43M+ embodied visual seek-and-follow samples from 770+ environments under both single-person and multi-person settings.}
  \label{benchoverview}
\end{figure*}
\subsection{Language-Guided Target Search}
Prior work has largely studied language-conditioned target search in neighboring embodied settings, particularly in object-goal navigation~\cite{chaplot2020object,garg2025objectreact,yadav2023ovrl,shah2023lm}. In these settings, the agent must ground natural language into embodied exploration in order to localize a target. However, these tasks are largely object-centric and do not truly address the retrieval of a specific person. More recently, text-driven target person retrieval has begun to receive attention, where language and visual cues are typically combined with map guidance and target-related prior information to localize the target person~\cite{park2023socrates,fung2025mllm}. Although these studies move closer to person-centered embodied search, they are still typically validated in specific environments and scenario instances, and they do not provide a unified and general benchmark for systematic evaluation. Their evaluation settings are often limited to narrow task formulations and do not sufficiently reflect the diversity and continuity of real embodied interaction. More importantly, they mainly focus on the retrieval stage itself and stop once the target person has been found, without considering the subsequent requirement of persistent human following. Consequently, they do not address how an agent should transition from target search to target-centered following, nor how it should maintain coherent behavior when the target is temporarily lost or reappears in the scene. As a result, existing work does not capture the full embodied interaction process in which an agent must first identify and locate a language-described target person and then continue to follow that person over time. This limitation makes it difficult to systematically study unified policies that must couple semantic target grounding, behavioral transition, and long-horizon target-centered interaction within a single task formulation. This gap motivates a benchmark that jointly evaluates target search, behavioral transition, and sustained target-centered interaction within a unified embodied setting.

\subsection{Embodied Human Following}
Human following~\cite{zhong2024empowering,gupta2016novel,zhang2024uni,fung2025ldtrack,wang2025trackvla,liu2025trackvla++} investigates how an embodied agent continuously follows a dynamic target using different sensing modalities. Recent end-to-end VLA-based following models further demonstrate the effectiveness of directly mapping multimodal observations to actions. Uni-NaVid~\cite{zhang2024uni} introduces a vision-language-action (VLA) model that improves human-following capability through large-scale training in simulation. However, its reliance on a discrete action space limits adaptability in real-world environments. Building on this direction, TrackVLA~\cite{wang2025trackvla} advances this line of research by jointly modeling recognition and planning within a unified framework, achieving strong performance in real-world tracking tasks. TrackVLA++~\cite{liu2025trackvla++} further improves embodied visual tracking by incorporating explicit spatial reasoning and long-horizon target memory, leading to stronger robustness under severe occlusions and visually similar distractors. However, existing methods generally assume that the target is already visible at the beginning of the task. As a result, these methods mainly focus on maintaining continuous following after the target has been identified, rather than first searching for the target or handling the transition from search to follow. Overall, prior work treats person retrieval and human following as separate tasks, thereby overlooking a more practical real-world requirement in which an embodied agent must first find a language-specified target person and then continue to follow that person within a unified interaction process. To address this gap, we introduce a large-scale, diverse, and generalizable benchmark for the unified task of language-guided human seeking and persistent following.

\section{UESF-Bench Construction}
\label{sec:method}

\subsection{Data Collection}
Existing embodied benchmarks mainly focus on human following with an initially visible target, while largely overlooking the human-like target search process that is often required before following can begin in real-world scenarios. As a result, a benchmark for unified language-guided human seeking remains lacking.
To fill this gap, we construct a large-scale, diverse, and general benchmark, termed UESF-Bench, that unifies human seeking and following in a single continuous task setting. We collect a total of 1.43M embodied visual seek-and-follow samples for UESF-Bench.
Unlike existing benchmarks that typically treat these two tasks separately, UESF-Bench requires an agent to first seek a language-described target person who is initially out of camera view and then seamlessly transition to following that person in dynamic environments.
\subsection{UESF-Bench Simulation Environment}
 We build the simulation environment of UESF-Bench on Habitat 3.0~\cite{puig2023habitat}, which provides a ready-to-use engine for rendering, navigation, and collision handling in indoor scenes. Our environment construction pipeline consists of five stages: human avatar generation, appearance description generation, motion generation, scene recognition, and instruction generation. Specifically, we use the SMPL-X human model to create diverse humanoid avatars with randomized body shapes and UV texture maps from the ATLAS dataset ~\cite{liu2024texdreamer}. We then employ the vision-language model GPT-5.2~\cite{singh2025openai} to produce a one-sentence textual description for each avatar based on its visible appearance, and adopt MotionConverterSMPLX from Habitat 3.0 to generate realistic human motion sequences. To provide scene-aware context for language instruction generation, we further perform scene recognition by randomly sampling the main human’s position and using GPT-5.2~\cite{singh2025openai} to infer the room type from the corresponding views of the main human. Finally, based on the generated appearance description and recognized scene context, we use Gemini-3-Pro~\cite{ma2026safety} to produce the seek-follow instruction for each episode.

\subsection{Embodied Visual Seek-and-Follow Benchmark}
Based on our simulation environment, we construct the Unified Embodied Visual Seek-and-Follow Benchmark (UESF-Bench) for systematic evaluation of this task, as illustrated in Fig.~\ref{benchoverview}. We create 4859 humanoid avatars with diverse appearances and generate corresponding appearance descriptions for each of them. UESF-Bench is built from 777 of the 890 scenes available in HM3D~\cite{ramakrishnan2021habitat} and MP3D~\cite{chang2017matterport3d}; the remaining scenes are excluded because they do not support valid search-to-follow episode generation under our construction criteria.
UESF-Bench contains a total of 15905 episodes, which are partitioned into training, validation, and testing splits with no overlap in scenes. Specifically, the training split contains 11,310   episodes from 543 scenes, the validation split contains 1510 episodes from 77 unseen scenes, and the test split contains 3085 episodes from 157 unseen scenes. To evaluate model performance under different social complexities, UESF-Bench includes both single-person and multi-person scenarios. Overall, we collect 1.43M embodied visual seek-and-follow samples. Each sample consists of a navigation history (RGB sequence), a target description, and the corresponding expert trajectory.

\begin{figure*}[t]
  \centering
  \includegraphics[width=1\linewidth]{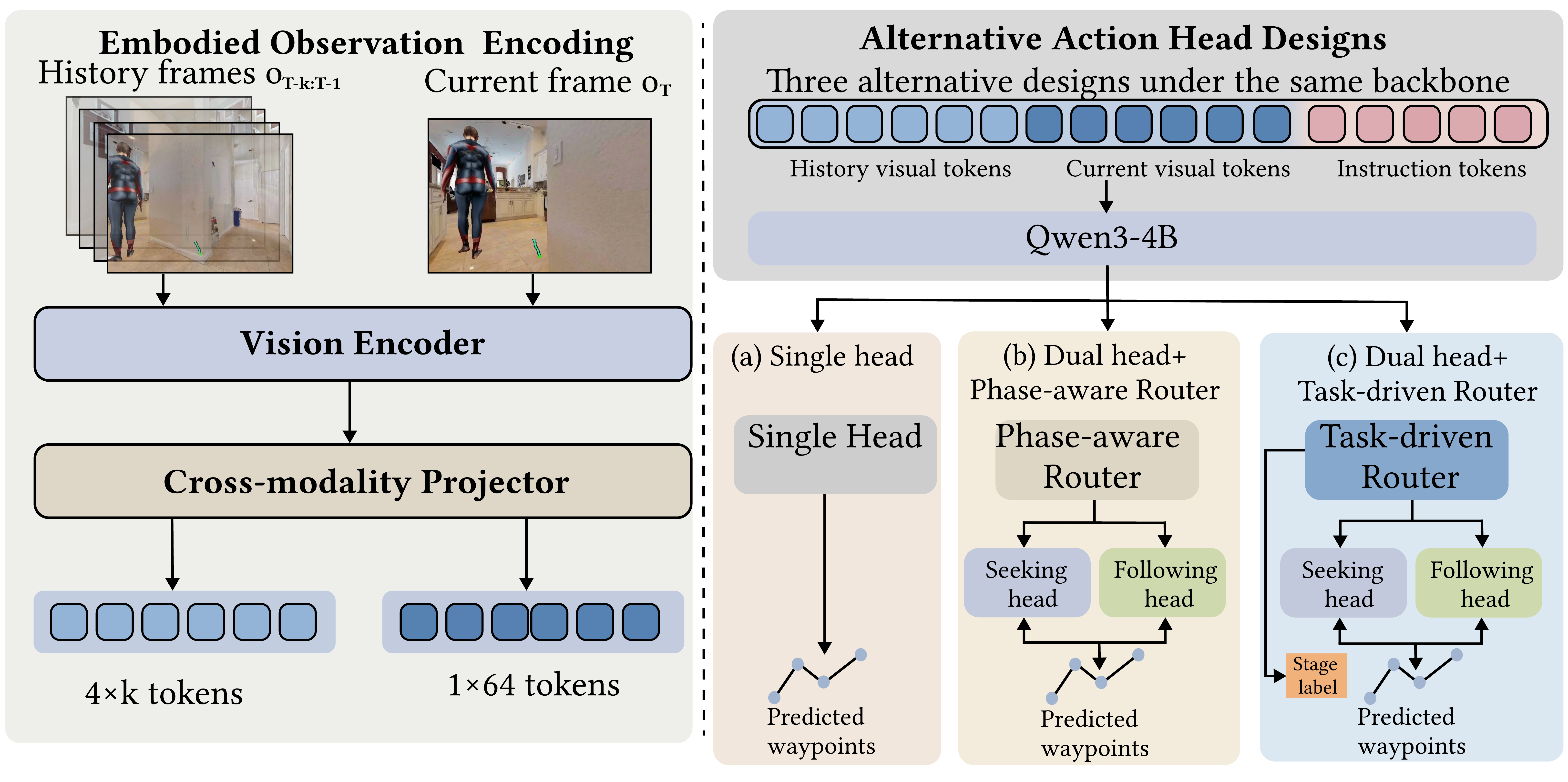}
  \caption{Architecture of SeekFollow-VLA. The model encodes historical and current observations into visual tokens, projects them into the language space, and combines them with instruction tokens in a shared Qwen3-4B backbone. On top of this backbone, we study three alternative action head designs for embodied seek-and-follow: a single head, dual heads with a phase-aware router, and dual heads with a task-driven router.}
  \label{seekvla}
\end{figure*}
\section{Methodology}
\subsection{Problem Formulation}

We formulate the embodied visual seek-and-follow task as a partially observable sequential decision problem. At each time step $t$, the agent receives a fused language instruction $l$ and an egocentric RGB observation history $O_t=\{o_1,\dots,o_t\}$, where the instruction specifies the appearance of a target person together with the seek-and-follow goal. The agent is required to transform the waypoints predicted by the model into control actions $a_t \in \mathcal{A}=\{v,\omega\}$, where $v$ and $\omega$ denote the linear and angular velocities of the agent, respectively.
Unlike conventional embodied visual tracking, the target person is typically outside the camera view at the beginning of the task.
As a result, the agent must first find a language-described target person who is initially out of camera view, and then transition seamlessly to following that person in dynamic environments.
Since no explicit signal is provided to indicate the transition between the seeking and following phases, the model must handle both tasks under a unified formulation. The task is considered successful only if the agent first seeks the correct target person and then continuously maintains an appropriate following distance (1--3 m) while facing toward the target.

\subsection{SeekFollow-VLA Overview}
%

 As shown in Fig.~\ref{seekvla}, SeekFollow-VLA adopts a pre-trained vision-language model, Qwen3-4B~\cite{ma2026safety}, as the backbone. For waypoint prediction, SeekFollow-VLA first encodes historical and current egocentric RGB observations into visual tokens, and then concatenates these tokens with the fused language instruction as the input to the multimodal backbone.

Based on the shared multimodal representation, SeekFollow-VLA predicts a sequence of future waypoints for agent control. To study how action prediction should be designed under the implicit transition between seeking and following, we instantiate three alternative action head designs under the same backbone: (1) a single shared action head, which directly maps the multimodal representation to waypoint predictions; (2) dual action heads with a phase-aware router, where separate seeking and following heads are combined through learned routing weights; and (3) dual action heads with a task-driven router, where the routing process is further guided by stage supervision to better distinguish search-oriented and follow-oriented behaviors.

This design allows us to systematically investigate how architectural inductive bias influences policy learning in embodied visual seek-and-follow. In particular, the task-driven dual-head variant explicitly addresses the mode confusion caused by the implicit stage boundary, enabling the model to switch more reliably between human seeking and human following behaviors within a unified framework.
\subsection{SeekFollow-VLA Architecture}
Given the egocentric RGB observation history \(O_T=\{o_1,\dots,o_T\}\), we encode each frame with a dual-tower visual encoder composed of DINO-V3~\cite{simeoni2025dinov3} and SigLIP~\cite{zhai2023sigmoid}. For each frame, the two vision towers extract spatial patch-level features under a unified input resolution of \(384\times384\). Let the resulting token grids be denoted as \(V^{\mathrm{dino}}\in\mathbb{R}^{P\times C_d}\) and \(V^{\mathrm{siglip}}\in\mathbb{R}^{P\times C_s}\),
\begin{equation}
V=\mathrm{Concat}\!\left(V^{\mathrm{dino}},V^{\mathrm{siglip}}\right)\in\mathbb{R}^{P\times C},
\end{equation}
where $P = H_p W_p$ denotes the number of aligned visual patches (set to 576); $C_s$ and $C_d$ denote the embedding dimensions of the SigLIP and DINO features, respectively; and $C = C_s + C_d$ denotes the total embedding dimension after feature concatenation.
To obtain compact multi-scale visual representations, we further apply a grid-pooling operation~\cite{zhang2024navid,zhang2024uni} to the fused patch tokens. Specifically, we generate two levels of visual tokens:
\begin{equation}
V^{\mathrm{fine}}=\mathrm{GridPool}(V,64),
\end{equation}

\begin{equation}
V^{\mathrm{coarse}}=\mathrm{GridPool}(V,4),
\end{equation}
where \(V^{\mathrm{fine}}\in\mathbb{R}^{64\times C}\) provides a fine-grained representation and \(V^{\mathrm{coarse}}\in\mathbb{R}^{4\times C}\) provides a compact coarse-grained representation.

To ensure computational efficiency during inference, the observation history  uses a sliding-window mechanism that retains  the most recent \(k\) historical frames together with the current frame, where T denotes the current time step and \(k=31\) in our experiments. Within this temporal window, we use the fine-grained tokens \(V_T^{\mathrm{fine}}\) for the latest observation and the coarse-grained tokens for historical observations to balance visual detail and token efficiency. Accordingly, the visual token sequence is organized as:
\begin{equation}
\mathcal{V}_T=\{V_{T-k}^{\mathrm{coarse}},\dots,V_{T-1}^{\mathrm{coarse}},V_T^{\mathrm{fine}}\}.
\end{equation}

Following established vision-language models (VLMs)~\cite{ma2026safety}, we use a cross-modality projector \(P(\cdot)\), implemented as a two-layer MLP, to project the visual token sequence \(\mathcal{V}_T\) into the latent space of the large language model:
\begin{equation}
E_T^{v}=P(\mathcal{V}_T).
\end{equation}

We concatenate the projected visual embeddings \(E_T^{v}\) with the text embeddings of the fused instruction and feed them into the multimodal backbone to obtain the output representation $E_T^{\mathrm{LLM}}$. This representation is then used as the input to the action prediction module for waypoint generation. Depending on the action head architecture, $E_T^{\mathrm{LLM}}$ is processed by a single action head, dual action heads with routing, or dual action heads with a task-driven router.

\subsection{Action Head Design}

A key challenge of the embodied visual seek-and-follow  task is that it unifies human seeking and human following within a single task, while the transition between the two stages is implicit. The objectives of its two stages are fundamentally different. During the seeking stage, the agent aims to quickly locate the target person described by language, whereas during the following stage, it must continuously track and follow that dynamic target.
Existing VLA designs with either a single action head or dual action heads may suffer from mode confusion under our setting. To systematically study how action prediction should be structured under implicit seek-to-follow transitions, we investigate three action head architectures.

\subsubsection{Single Action Head}
As the unified seek-and-follow task is formulated end-to-end, a straightforward formulation is to use a single shared action head that maps the multimodal representation $E_T^{\mathrm{LLM}}$  to the waypoint sequence. The action head is implemented as a three-layer MLP. Specifically, the input is first normalized by LayerNorm, then passed through two hidden linear layers with GELU activations, followed by a final linear projection to produce an output vector of dimension $M \times d_a$, where $M$ (set to 10) is the number of predicted waypoints and $d_a$ (set to 3) is the action dimension of each waypoint. A $\tanh$ function is applied to bound the output within $[-1,1]$. The resulting vector is then reshaped into $\hat{\mathbf{W}} \in \mathbb{R}^{M \times d_a}$. Each waypoint \(w_i= (x_i, y_i, \theta_i)_{i=1}^{M}\) consists of the agent's position \( x \), \( y \) in the plane, and \( \theta \), which represents the heading or orientation of the agent,

\begin{equation}
\hat{\mathbf{W}}_T = \mathrm{SingleActionHead}\!\left(\mathbf{E}_T^{\mathrm{LLM}}\right).
\end{equation}

The single action head is simple and parameter-efficient, but it does not explicitly account for the behavioral differences between seeking and following.
As a result, modeling both behaviors with a single action head may lead to mode confusion. This observation motivates our dual-head design with adaptive routing.

\subsubsection{Dual Action Heads with Phase-Aware Router}
To better capture the behavioral differences between seeking and following, we replace the shared action head with two separate heads: a seeking head and a following head. The two heads share the same architecture as the single action head, but maintain separate parameters for seeking and following. In this way, seeking and following are modeled by different prediction branches,
\begin{equation}
\hat{\mathbf{W}}_T^{\mathrm{seek}} = \mathrm{SeekingActionHead}\!\left(\mathbf{E}_T^{\mathrm{LLM}}\right),
\end{equation}
\begin{equation}
\hat{\mathbf{W}}_T^{\mathrm{follow}} = \mathrm{FollowingActionHead}\!\left(\mathbf{E}_T^{\mathrm{LLM}}\right).
\end{equation}
Furthermore, since UESF is a continuous task without explicit phase boundaries, we introduce a Phase-Aware Router to adaptively determine which head should contribute more to the final prediction. The Phase-Aware Router is implemented as a three-layer MLP that takes the multimodal representation $E_T^{\mathrm{LLM}}$ 
as input and outputs two routing weights for the seeking and following heads. These scores are normalized by a softmax function to obtain routing weights, which are then used to combine the outputs of the two action heads,
\begin{equation}
\boldsymbol{\alpha}_T = \mathrm{PhaseAwareRouter}\!\left(\mathbf{E}_T^{\mathrm{LLM}}\right),
\end{equation}
where $\boldsymbol{\alpha}_T = [\alpha_T^{\mathrm{seek}}, \alpha_T^{\mathrm{follow}}]$ denotes the routing weights assigned to the two heads. Using these routing weights, the final waypoint sequence is obtained by fusing the outputs of the two heads,

\begin{equation}
\hat{\mathbf{W}}_T
=
\alpha_T^{\mathrm{seek}} \hat{\mathbf{W}}_T^{\mathrm{seek}}
+
\alpha_T^{\mathrm{follow}} \hat{\mathbf{W}}_T^{\mathrm{follow}},
\end{equation}
where $\hat{\mathbf{W}}_T$ denotes the final predicted waypoint sequence obtained by fusing the outputs of the seeking and following heads.

Compared with the single-head formulation, this design explicitly decouples the two behavior modes into separate prediction branches, allowing each head to specialize in predicting a more coherent trajectory. At the same time, this design ensures that the two behaviors remain within the same unified task framework, enabling seamless transitions between them.
Nevertheless, the router is still optimized solely through the final waypoint prediction objective, without explicit supervision on which behavior mode should dominate at each time step. Under partial observability, this lack of supervision may lead to ambiguous routing decisions.

\subsubsection{Dual Action Heads with a Task-Driven Router}

To improve the adaptability of the model, we propose a Task-Driven Router. Like the Phase-Aware Router, it is implemented as a three-layer MLP that incorporates task-specific cues to guide the routing process.
This task-driven router enables the model to more effectively determine which head should dominate at each time step, ensuring a more accurate and stage-aware prediction.

The two action heads remain the same as in the previous design:
\begin{equation}
\hat{\mathbf{W}}_T^{\mathrm{seek}} = \mathrm{SeekingActionHead}\!\left(\mathbf{E}_T^{\mathrm{LLM}}\right),
\end{equation}
\begin{equation}
\hat{\mathbf{W}}_T^{\mathrm{follow}} = \mathrm{FollowingActionHead}\!\left(\mathbf{E}_T^{\mathrm{LLM}}\right).
\end{equation}

Different from the Phase-Aware Router, the Task-Driven Router outputs both routing weights and routing logits:
\begin{equation}
(\boldsymbol{\alpha}_T,\mathbf{g}_T)=\mathrm{TaskDrivenRouter}\!\left(E_T^{\mathrm{LLM}}\right),
\end{equation}
where $\boldsymbol{\alpha}_T=[\alpha_T^{\mathrm{seek}},\alpha_T^{\mathrm{follow}}]$ denotes the routing weights assigned to the two heads, and $\mathbf{g}_T=[g_T^{\mathrm{seek}},g_T^{\mathrm{follow}}]$ denotes the corresponding task-type logits. The final prediction is given by,
\begin{equation}
\hat{\mathbf{W}}_T
=
\alpha_T^{\mathrm{seek}} \hat{\mathbf{W}}_T^{\mathrm{seek}}
+
\alpha_T^{\mathrm{follow}} \hat{\mathbf{W}}_T^{\mathrm{follow}}.
\end{equation}

\subsection{Training Objective}

For all model variants, the primary training objective is waypoint prediction.
Given the predicted waypoint sequence $\hat{\mathbf{W}} \in \mathbb{R}^{B \times M \times D}$ and the ground-truth waypoint sequence $\mathbf{W} \in \mathbb{R}^{B \times M \times D}$, we compute a masked mean squared error over the selected waypoints:
\begin{equation}
\mathcal{L}_{\mathrm{wp}}
=
\frac{1}{|\Omega|}
\sum_{(b,m,d)\in\Omega}
\left(
\hat{\mathbf{W}}_{b,m,d}-\mathbf{W}_{b,m,d}
\right)^2,
\end{equation}
where $\Omega$ denotes the set of valid waypoint entries selected by the waypoint mask, $B$ is the batch size, $M$ is the number of predicted waypoints, and $D$ is the waypoint dimension. If no valid waypoint is selected, the loss is set to zero.

For the Single Action Head and Dual Action Heads with Router variants, the training objective is defined as follows:
\begin{equation}
\mathcal{L}
=
\mathcal{L}_{\mathrm{wp}}.
\end{equation}

For the \textbf{Dual Action Heads with a Task-Driven Router} model variant, we further introduce an auxiliary task supervision on the predicted task logits. Specifically, the Task-Driven Router outputs both routing weights and task logits, where the routing weights are used for waypoint fusion and the task logits are supervised by a task-related label $z_T \in \{0,1\}$. Here, \(z_T\) is the ground-truth stage label, where \(z_T=0\) indicates the seeking stage and \(z_T=1\) indicates the following stage. The label is defined by task progression. Specifically, \(z_T\) remains 0 until the agent first successfully finds the target person, where success is defined as the target being within \(1.7\,\mathrm{m}\) and the agent facing the target. After this first successful discovery, \(z_T\) remains 1 for all subsequent time steps. An auxiliary stage-aware loss is defined as a cross-entropy loss:
\begin{equation}
\mathcal{L}_{\mathrm{stage}}
=
\mathrm{CrossEntropy}(\mathbf{g}_T, z_T),
\end{equation}
where $\mathbf{g}_T=[g_T^{\mathrm{seek}}, g_T^{\mathrm{follow}}]$ denotes the predicted task logits, and $z_T$ indicates the ground-truth behavior mode at the current step.

The final objective for the task-driven variant is defined as the weighted sum of the waypoint prediction loss and the auxiliary stage-aware loss,
\begin{equation}
\mathcal{L}
=
\beta_{\mathrm{nav}} \mathcal{L}_{\mathrm{wp}}
+
\mathcal{L}_{\mathrm{stage}},
\end{equation}
where, \(\beta_{\mathrm{nav}}\) controls the relative weight of the waypoint prediction loss. In all experiments, we set \(\beta_{\mathrm{nav}} = 10\). In this way, the waypoint regression term optimizes trajectory prediction, while the auxiliary gate loss explicitly encourages correct task-aware routing between the seeking and following modes.

\section{Experiments}
\label{sec:experiments}

\subsection{Experimental Setup}
We follow the architecture described in Sec. 4.2. Unless otherwise specified, the three compared model variants differ only in the action prediction head. For both the single-person and multi-person settings, we construct training, validation, and test splits. All models are trained on the training split, selected based on validation performance, and finally evaluated on the corresponding test split. Training is conducted on five NVIDIA H100 GPUs using distributed mixed-precision. We use an initial learning rate of $2 \times 10^{-5}$, a batch size of 22, a waypoint horizon of 10, and a temporal history length of 31. Validation is performed every 1000 training steps. We set the maximum number of training epochs to 10 and apply early stopping based on validation performance. In practice, the best checkpoints are obtained within 2--3 epochs.

\subsection{Metrics}
To comprehensively evaluate performance on the embodied visual seek-and-follow task, we adopt four metrics covering task success, safety, following quality, and search efficiency. These metrics are defined as follows:
\begin{itemize}
    \item \textbf{Task Success Rate (TSR):} the ratio of episodes that successfully complete the overall seek-and-follow task to the total number of episodes.
    
    \item \textbf{Collision Rate (CR)~\cite{wang2025trackvla}:} the ratio of episodes with at least one robot-human collision during the following phase to the total number of episodes.
    
    \item \textbf{Following Ratio (FR)~\cite{wang2025trackvla}:} the ratio of successful following steps to the total number of following-phase steps.
    
    \item \textbf{Search SPL~\cite{anderson2018vision}:} the success-weighted path length in the seeking phase, reflecting both search success and path efficiency, with values ranging from 0 to 1.
\end{itemize}

\subsection{Single-Person: Main Comparison and Ablation}

We first evaluate the three action head architectures in the single-person setting. Specifically, we compare the Single Action Head (SingleHead), Dual Action Heads with Phase-Aware Router (DualHead-PA), and Dual Action Heads with a Task-Driven Router (DualHead-TD) on the single-person test split using TSR, CR, FR, and Search SPL as evaluation metrics. As shown in Table ~\ref{tab:single}, the quantitative results of the three methods are compared under the single-person setting. 

In the single-person setting, DualHead-TD achieves the highest TSR of 0.35, compared with 0.04 for SingleHead and 0.05 for DualHead-PA. It also attains the highest FR of 0.92, whereas both SingleHead and DualHead-PA remain at 0.85. In addition, DualHead-TD achieves the best Search SPL of 0.53, clearly outperforming SingleHead (0.30) and DualHead-PA (0.27). These results indicate that task-driven routing effectively improves both search efficiency and following stability, leading to stronger end-to-end performance in the unified seek-and-follow task.

By contrast, SingleHead and DualHead-PA perform similarly in the single-person setting, especially in terms of TSR, where both remain at a very low level. This suggests that neither using a single shared action head nor introducing a dual-head structure without explicit task-driven supervision is sufficient to address the core challenges of this task. Specifically, the model must not only handle the implicit transition between seeking and following, but also resolve the behavioral ambiguity that arises when the target temporarily leaves the field of view during the following stage.


\begin{table}[t]
  \centering
  \caption{%
    \textbf{Single-person results} on UESF-Bench test set.
    TSR: Task Success Rate; CR: Collision Rate;
    FR: Following Ratio; SPL: Search SPL.
  }
  \label{tab:single}
  \begin{tabular*}{\columnwidth}{@{\extracolsep{\fill}}ccccc@{}}
    \toprule
    \textbf{Method} & \textbf{TSR} & \textbf{CR} & \textbf{FR}  & \textbf{SPL} \\
    \midrule
    SingleHead                & 0.04 & 0.07 & 0.85 & 0.30 \\
    DualHead-PA               & 0.05 & 0.06 & 0.85 & 0.27 \\
    \textbf{DualHead-TD} & \textbf{0.35} & \textbf{0.32} & \textbf{0.92} & \textbf{0.53} \\
    \bottomrule
  \end{tabular*}
\end{table}

\subsection{Multi-Person: Main Comparison and Ablation}
To evaluate the three action head architectures in the more challenging multi-person setting, we place 2--7 distractor persons in each scene, with trajectories that frequently intersect with that of the target person. This setting not only makes target identification more difficult during the seeking stage, but also introduces persistent identity ambiguity during following. As a result, the task becomes substantially more challenging than in the single-person setting. Table~\ref{tab:multi} summarizes the quantitative results on the multi-person test split.

Compared with the single-person setting, performance on most metrics declines in the multi-person setting, indicating that distractor interference substantially increases the difficulty of embodied visual seek-and-follow.
For DualHead-TD, TSR drops from 0.35 to 0.20, while FR decreases from 0.92 to 0.82. For SingleHead, FR also decreases from 0.85 to 0.73. 
Although the metric changes are not entirely consistent across models, the overall trend shows that the multi-person setting is clearly more challenging. When multiple distractors are present, the model must not only identify the target during seeking, but also keep verifying whether the currently observed person is the intended one during following. This becomes even harder under occlusion, trajectory intersections, and close parallel movement. These results suggest that multi-person interference is not simply a matter of increased scene complexity; it also introduces additional challenges for target identification, stage transition, and sustained following.

Even under this more challenging setting, DualHead-TD achieves the highest TSR of 0.20, whereas both SingleHead and DualHead-PA remain at 0.04. This result indicates that explicit task-driven routing is particularly important for end-to-end task success in the presence of multiple distractors. In terms of Search SPL, DualHead-TD achieves 0.55, again significantly outperforming the other two methods. This result indicates that it not only identifies the correct target more reliably, but also performs the seeking process more efficiently. It also suggests that, when multiple distractors are present, selecting an appropriate behavior mode according to the current task stage is more important than merely improving local capability within a single stage.
For FR, DualHead-PA achieves the highest value of 0.85, while DualHead-TD reaches 0.82. Although DualHead-TD does not achieve the best FR, its clear advantages in TSR and Search SPL better reflect its overall suitability for the unified seek-and-follow task.

In both the single-person and multi-person settings, DualHead-TD consistently achieves the best TSR and Search SPL, showing that task-driven routing effectively reduces mode confusion between seeking and following under implicit stage transitions. In contrast, SingleHead entangles the two behaviors within a shared action head, making it difficult to learn clear stage-specific policies. Although DualHead-PA adopts a dual-head structure, it lacks explicit task-driven supervision and therefore cannot reliably separate the demands of seeking and following. While DualHead-TD also exhibits a higher collision rate, this trend should be interpreted together with its stronger task success and search performance rather than viewed in isolation. Overall, these results validate the effectiveness of the proposed task-driven dual-head design for the unified seek-and-follow task.

\begin{table}[t]
  \centering
  \caption{%
    \textbf{Multi-person results} on UESF-Bench test set.
    Same metrics as Table~\ref{tab:single}.
  }
  \label{tab:multi}
  \begin{tabular*}{\columnwidth}{@{\extracolsep{\fill}}ccccc@{}}
    \toprule
    \textbf{Method} & \textbf{TSR} & \textbf{CR} & \textbf{FR} & \textbf{SPL} \\
    \midrule
    SingleHead                & 0.04 & 0.06 & 0.73 & 0.31 \\
    DualHead-PA               & 0.04 & 0.06 & 0.85 & 0.26 \\
    \textbf{DualHead-TD} & \textbf{0.20} & \textbf{0.33} & \textbf{0.82} & \textbf{0.55} \\
    \bottomrule
  \end{tabular*}
\end{table}

\subsection{Routing Dynamics Analysis}
\begin{figure}[t]
  \centering
  \includegraphics[width=\linewidth]{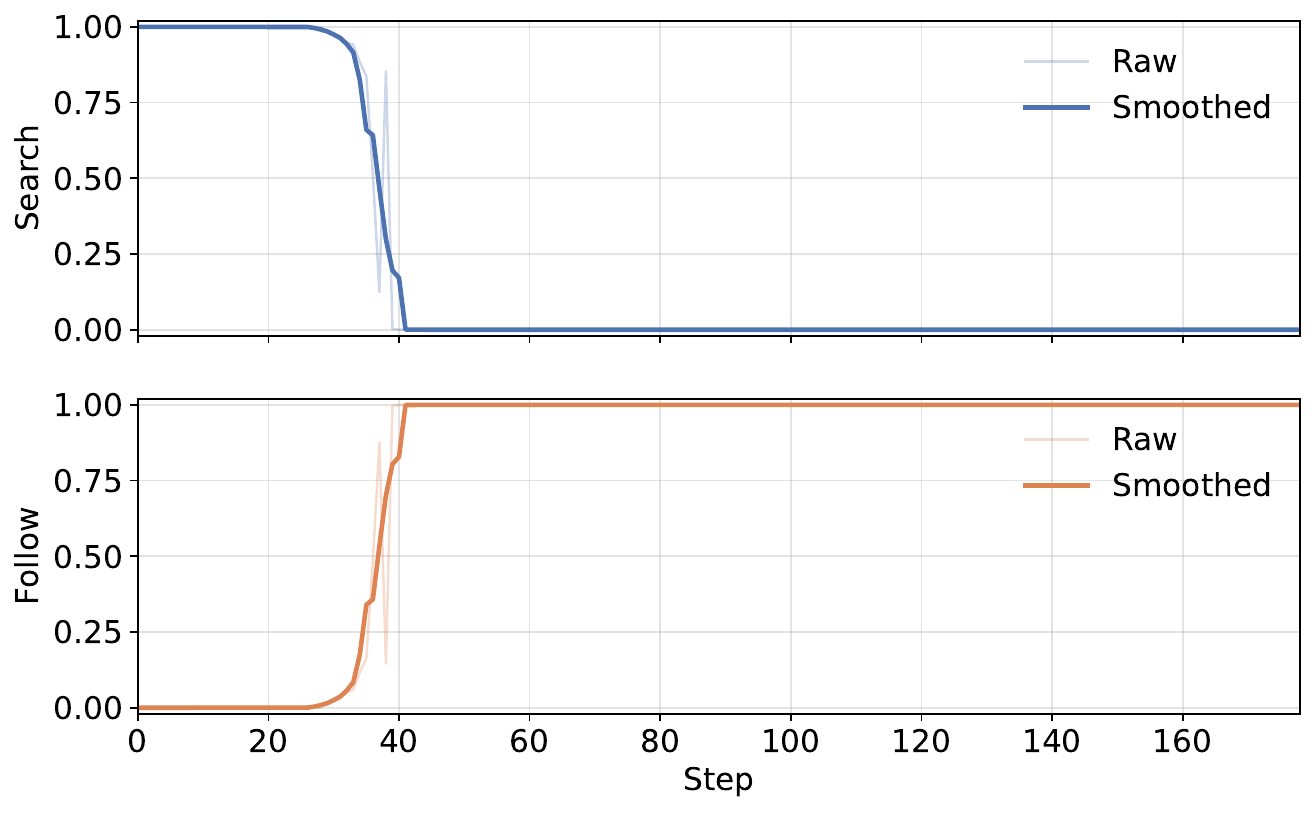}
  \caption{%
    \textbf{Router weights $p_\text{seek}$ and $p_\text{follow}$ over episode time.}
  }
  \label{fig:gate_vis}
\end{figure}


Fig.~\ref{fig:gate_vis} visualizes the router weights of the task-driven variant in a representative episode. A clear two-stage routing pattern can be observed. During the early part of the episode, the router assigns nearly all weight to the search head, indicating that the model primarily performs exploratory target seeking. Although slight oscillations appear in intermediate steps, the routing remains consistently dominated by the search branch. Once the target is found, the routing undergoes a sharp transition: the search weight quickly collapses to nearly zero, while the follow weight rises to nearly one and remains stable thereafter. Such behavior shows that the router learns an explicit functional separation between search and follow behaviors, despite the overall task being defined under a unified instruction without an explicit phase boundary at inference time. This supports our claim that task-driven routing improves behavioral allocation and helps the model handle implicit seek-to-follow transitions more reliably. Additional experimental results and visualizations are provided in the supplementary material.


\section{Conclusion}
\label{sec:conclusion}
In this work, we introduce embodied visual seek-and-follow as a unified task in which an agent must first locate a target person described by language and then persistently follow that person under an implicit phase boundary. This setting extends beyond conventional human-following tasks by jointly requiring semantic-guided target search, reliable behavior switching and recovery when the target appears or is temporarily lost, and delayed identity grounding under distractors. To enable systematic study of this problem, we construct UESF-Bench, a large-scale and diverse benchmark that covers both single-person and multi-person scenarios. We further propose SeekFollow-VLA, a vision-language-action framework with a task-driven routing design for latent phase inference and behavior coordination across the seeking and following stages. Experimental results show that SeekFollow-VLA consistently outperforms the compared baselines across different settings, with especially clear advantages in more challenging multi-person environments.


\bibliographystyle{ACM-Reference-Format}
\bibliography{sample-base}










\end{document}